\pdfoutput=1
\documentclass[11pt]{article}

\usepackage[preprint]{acl}
\usepackage{times}
\usepackage{latexsym}
\usepackage[T1]{fontenc}
\usepackage[utf8]{inputenc}
\usepackage{microtype}
\usepackage{inconsolata}
\usepackage{graphicx}
\usepackage{amsmath}
\usepackage{amssymb}
\usepackage{algorithm}
\usepackage{algpseudocode}
\usepackage{float}
\usepackage{booktabs}
\usepackage{multirow}

\title{REED: Post-Training Representation Editing for Cross-Domain \\ Linguistic Steganalysis}

\author{
  Ruohan Lei\textsuperscript{1},
  Jianxin Gao\textsuperscript{1},
  Wanli Peng\textsuperscript{1,*},
  Huimin Pei\textsuperscript{2}
  \\
  \textsuperscript{1}China Agricultural University, Beijing, China
  \\
  \textsuperscript{2}Jiangsu Normal University, Xuzhou, China
  \\
  \texttt{\{lrh07,jxgao,wlpeng\}@cau.edu.cn},
  \texttt{hmpei@jsnu.edu.cn}
  \\
  \textsuperscript{*}Corresponding author.
}

\begin{document}
\maketitle

\begin{abstract}
In real-world scenarios of linguistic steganalysis, tested texts usually come from unseen domains with different vocabularies, topics, writing styles, and steganographic generation patterns, which can significantly degrade the detection performance. 
Although existing cross-domain steganalysis methods can effectively alleviate this problem through distribution alignment, domain-invariant feature learning, etc., the detection performance is not satisfactory.
In this paper, we propose a post-training representation editing method for cross-domain linguistic steganalysis. 
Specifically, the detector is first trained on source-domain data, and then the feature extractor and classifier are kept frozen, and the intermediate representations are deterministically edited before classification. 
For domain adaptation, we construct a domain-offset vector from marginal source and target representations. 
For domain generalization, we derive a source-domain cover-to-stego direction to guide sample-specific editing. 
Experimental results show that compared with the advanced methods, the proposed method can achieve high cross-domain detection performance, especially in terms of F1-score, while requiring no architecture modification or parameter updates after source-domain training.
\end{abstract}

\section{Introduction}

Linguistic steganography embeds secret messages into natural text, enabling covert communication over public channels \citep{anderson1998limits}. 
Recent generation-based methods, driven by neural language models, can produce fluent stego texts with high embedding capacity and strong imperceptibility \citep{ziegler2019neural,zhang2021provably,zhou2022linguistic}. 
While this progress improves the practicality of steganography, it also increases the risk of abusing public text channels to transmit hidden information \citep{mazurczyk2018information}. 
Linguistic steganalysis therefore aims to distinguish stego texts from normal cover texts and has become an important countermeasure.

Existing steganalysis methods mainly rely on deep neural detectors. 
including convolutional neural networks (CNNs), recurrent neural networks (RNNs), graph-based modeling, hierarchical learning, social-network-oriented detection, and attention-based feature fusion\citep{wen2019cnn,yang2019tsrnn,niu2019hybrid,peng2021realtime,wu2021linguistic,xue2022effective,yang2023linguistic,peng2023text}. 
These methods achieve promising results under the independent identically distributed (i.i.d.) assumption. 
In practice, however, this assumption is often violated. 
Suspicious texts may come from domains different from the training data and may differ in vocabulary, topic, length, and writing style, which can severely degrade the performance of a source-trained detector \citep{xue2024sanet,wu2024ttals}.

Similar to other classification tasks, cross-domain linguistic steganalysis focuses on two mismatch problems, containing the domain adaptation and the domain generalization \citep{xue2022mda,xue2024sanet,luo2025pdts,yang2025cada,wu2024ttals,wang2021tent}. 
For domain adaptation, learning domain-invariant features and adversarial learning are the commonly used paradigms.
For domain generation, the effectiveness of the test-time generalization has been explored.
Although the above methods improve the detection performance, these methods cannot achieve satisfactory performance, especially for domain generalization. 
The main reason is that once a detector is trained on the source domain, its classifier fixes the decision boundary in the learned representation space, and the target-domain representations are shifted away from the source distribution and become mismatched with this boundary \citep{xue2024sanet,yang2025cada}. 
Moreover, they usually require additional optimization objectives, target-domain fine-tuning, or test-time parameter updates, which require high computational cost.

This above analysis leads to an interesting question: \textbf{\textit{instead of parameter modification on target domain data, can we directly edit the representations before classification?}}

In this paper, we propose a lightweight representation editing framework, called \textbf{\textit{REED}}, for cross-domain linguistic steganalysis. 
The steganographic detector is first trained on source domain data, and then each test representation is edited by a task-dependent vector before being passed to the classifier. 
For domain adaptation, the vector is estimated from the marginal representation offset between the source and target domains. 
For domain generalization, where target samples are unavailable before testing, the editing direction is derived from the source-domain cover-stego representation difference. 
The proposed REED operates directly in the learned feature space, without modifying the architecture, introducing extra training losses, or updating model parameters after source-domain training. 
Experimental results show that the REED improves cross-domain detection performance for both domain adaptation and generalization tasks.

\section{Related Work}
\subsection{Linguistic Steganography}

Linguistic steganography aims to embed secret information into natural text while hiding the existence of communication \citep{anderson1998limits,ziegler2019neural}. Early studies mainly follow the modification-based paradigm, where secret bits are embedded by modifying an existing cover text, such as through synonym substitution or format changes \citep{qi2013secure,xiang2014linguistic,shirali2008text}. Since natural text has limited redundancy, these methods often suffer from restricted embedding capacity and may affect text naturalness.

With the development of neural language models, generation-based linguistic steganography has become a major direction \citep{fang2017generating,yang2019rnnstega,ziegler2019neural,zhang2021provably}. Instead of modifying a given text, these methods generate stego text directly by controlling word or token selection during decoding. Early neural methods use language models to map secret bits to candidate words, while later coding-based methods further exploit the probability distribution of the language model. For example, Huffman Coding assigns variable-length codes according to token probabilities, and Arithmetic Coding maps secret bits to probability intervals to improve embedding efficiency \citep{yang2018automatically,ziegler2019neural}. Adaptive Dynamic Grouping further partitions candidate tokens according to their probabilities, reducing the distortion introduced during generation \citep{zhang2021provably}. More recent methods such as Meteor and iMEC study stronger security guarantees for generative steganography \citep{kaptchuk2021meteor,witt2023perfectly}. These advances make stego text increasingly fluent and difficult to detect, which in turn raises the need for reliable linguistic steganalysis.

\subsection{Linguistic Steganalysis}

To counter the development of linguistic steganography, existing linguistic steganalysis methods mainly employ deep learning models to learn discriminative features between cover and stego texts. 
Early neural approaches use CNNs, RNNs, or hybrid architectures to capture textual patterns for cover-stego classification \citep{wen2019cnn,yang2019tsrnn,niu2019hybrid}. 
Recent work has further improved feature learning for linguistic steganalysis. 
Fine-tuned pre-trained language models are used to capture more subtle inconsistencies in generated text, while graph-based modeling, hierarchical learning, and attention mechanisms have also been explored to enhance steganalysis representations \citep{devlin2019bert,peng2021realtime,wu2021linguistic,xue2022effective,peng2023text}. 
In addition, some studies consider more realistic detection scenarios, such as linguistic stegananalysis on social networks \citep{yang2023linguistic}.


\subsection{Cross-Domain Linguistic Steganalysis}

To alleviate the performance degradation caused by domain mismatch, existing studies mainly resort to domain adaptation to improve the model's transferability to the target domain \citep{xue2022mda,xue2024sanet,luo2025pdts,yang2025cada}. MDA \citep{xue2022mda} first introduces domain adaptation and transductive learning into text steganalysis, where the distributions of source and target features are aligned with Maximum Mean Discrepancy (MMD) \citep{gretton2012kernel}. SANet \citep{xue2024sanet} further designs a Steganographic Domain Distance Metric (SDDM) and an adaptive weight selection network to measure source--target discrepancy and enhance domain-invariant feature extraction. PDTS \citep{luo2025pdts} adopts a two-stage framework, which first pre-trains the model on labeled source-domain data and then fine-tunes it with pseudo-labels from the target domain to capture target-domain-specific features. More recently, CADA \citep{yang2025cada} introduces class-aware adversarial pre-training and class-aware fine-tuning, where class-aware alignment and class-balanced pseudo-labels are used to improve both transferability and discrimination.

Beyond domain adaptation, domain generalization considers a more restrictive setting, where the model is expected to generalize to unseen target domains without accessing target-domain data during training \citep{zhou2022domain}. In this context, test-time adaptation has been explored as a practical way to improve out-of-distribution prediction by adapting a trained model using test data only \citep{sun2020test,wang2021tent,niu2022efficient}. TTALS \citep{wu2024ttals} brings this idea into cross-domain linguistic steganalysis and updates the model during testing by minimizing the prediction entropy \citep{wang2021tent}.

Overall, existing cross-domain methods usually adapt the model through additional alignment losses, pseudo-label-based fine-tuning, or test-time parameter updates \citep{xue2024sanet,yang2025cada,wu2024ttals}. In contrast, our method keeps the source-trained model fixed and directly edits intermediate representations after training. It therefore provides a lightweight post-training alternative that operates in feature space without changing the model architecture or updating model parameters.

\section{Method}

\subsection{Problem Formulation}

In domain adaptation, labeled source-domain texts and unlabeled target-domain texts are available \citep{xue2022mda,xue2024sanet,yang2025cada}. We denote them as
\begin{equation}
\mathcal{D}_s=\{(x_i^s,y_i^s)\}_{i=1}^{N_s}, \quad
\mathcal{D}_t=\{x_j^t\}_{j=1}^{N_t},
\end{equation}
where $x_i^s$ and $x_j^t$ are source- and target-domain texts, respectively, and $y_i^s\in\{0,1\}$ denotes the source label, with $0$ for cover and $1$ for stego. Target-domain labels are unavailable during training.

In domain generalization, only the labeled source-domain dataset $\mathcal{D}_s$ is available before testing, while target-domain texts and labels are inaccessible \citep{wu2024ttals}. This setting is more restrictive, since the detector must generalize to unseen domains without observing target-domain samples in advance.

We denote the source-trained steganalysis model as $h(x)=g_\phi(f_\theta(x))$, where $f_\theta$ is the feature extractor and $g_\phi$ is the classifier. The intermediate representation produced by $f_\theta$ is denoted by $z$.

\subsection{Vector-Guided Representation Editing}

\begin{figure*}[t]
    \centering
    \includegraphics[width=0.95\textwidth]{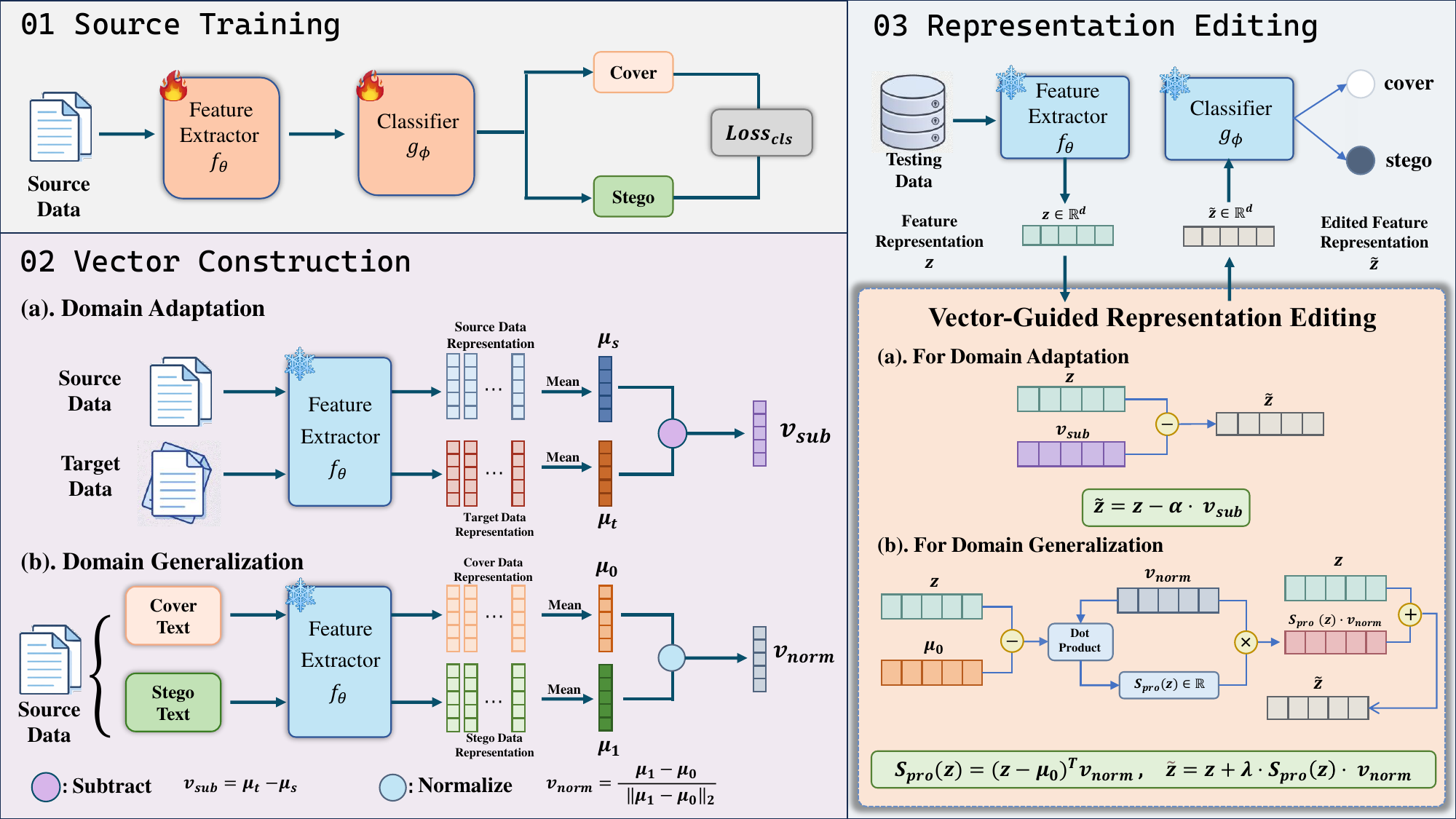}
    \caption{Overview of the proposed REED framework. After source-domain training, intermediate representations are edited by task-specific vectors before classification.}
    \label{fig:method_overview}
\end{figure*}

The overall framework of the proposed method is illustrated in Figure~\ref{fig:method_overview}.

We propose a vector-guided representation editing framework for cross-domain linguistic steganalysis. Unlike methods that introduce auxiliary adaptation losses \citep{xue2022mda,xue2024sanet,yang2025cada} or update model parameters at test time \citep{wu2024ttals,wang2021tent}, our method keeps the source-trained model frozen and edits intermediate representations before classification.

The framework contains three stages. First, a base steganalysis model is trained on labeled source-domain data following the adopted backbone, including its feature extractor, classifier, trainable-parameter setting, and training protocol. No target-domain samples or extra adaptation losses are used in this stage. Second, after source-domain training, all parameters are frozen and editing vectors are constructed from intermediate representations. In domain adaptation, the vector is estimated from source and unlabeled target representations; in domain generalization, it is derived from source-domain cover and stego representations. Third, during testing, each input representation is edited before being passed to the frozen classifier:
\begin{align}
 z &= f_\theta(x), \quad z\in\mathbb{R}^{d}, \label{eq:representation} \\
 \tilde{z} &= z+\Delta z, \label{eq:general_adjustment} \\
 \hat{y} &= \arg\max_{c\in\{0,1\}} g_\phi(\tilde{z})_c . \label{eq:prediction}
\end{align}
Here, $\Delta z$ is constructed differently under domain adaptation and domain generalization, as described below.

\subsection{Domain-Offset Representation Editing for Domain Adaptation}

In the domain adaptation setting, unlabeled target-domain texts allow us to estimate the source-to-target shift in the learned representation space. We approximate this shift with a global domain-offset vector. Let
\begin{equation}
\mathcal{D}_{s}^{\mathrm{est}}=\{x_i^s\}_{i=1}^{M_s}, \quad
\mathcal{D}_{t}^{\mathrm{est}}=\{x_j^t\}_{j=1}^{M_t}
\end{equation}
denote the source- and target-domain texts used for vector estimation, respectively. Using the frozen feature extractor, we compute their mean representations as
\begin{equation}
\mu_s=\frac{1}{M_s}\sum_{i=1}^{M_s}f_\theta(x_i^s), \quad
\mu_t=\frac{1}{M_t}\sum_{j=1}^{M_t}f_\theta(x_j^t).
\end{equation}
We define the subtraction vector as
\begin{equation}
v_{\mathrm{sub}}=\mu_t-\mu_s, \quad v_{\mathrm{sub}}\in\mathbb{R}^{d}.
\end{equation}
For a target-domain test sample $x$ with representation $z=f_\theta(x)$, we edit its representation by
\begin{equation}
\tilde{z}=z-\alpha v_{\mathrm{sub}},
\end{equation}
where $\alpha$ controls the editing strength. Unless otherwise specified, we set $\alpha=1$ in our experiments. When target representations are globally shifted from the source distribution, subtracting $v_{\mathrm{sub}}$ provides a direct editing toward the representation space learned by the source-trained classifier.

\subsection{Steganographic-Direction Representation Editing for Domain Generalization}

In the domain generalization setting, target-domain samples are unavailable before testing, so the source-to-target offset cannot be estimated. Instead, we derive a cover-to-stego direction from source-domain representations to guide the editing of unseen-domain representations.

Let
\begin{equation}
\mathcal{D}_s^0=\{x_i^0\}_{i=1}^{N_0}, \quad
\mathcal{D}_s^1=\{x_i^1\}_{i=1}^{N_1}
\end{equation}
denote source-domain cover and stego texts, respectively. Using the frozen feature extractor, we compute their mean representations as
\begin{equation}
\mu_0=\frac{1}{N_0}\sum_{i=1}^{N_0}f_\theta(x_i^0), \quad
\mu_1=\frac{1}{N_1}\sum_{i=1}^{N_1}f_\theta(x_i^1).
\end{equation}
The normalized cover--stego direction is defined as
\begin{equation}
 v_{\mathrm{norm}}=\frac{\mu_1-\mu_0}{\|\mu_1-\mu_0\|_2}, \quad
 v_{\mathrm{norm}}\in\mathbb{R}^{d}.
\end{equation}
For a test sample $x$, its representation $z$ is projected onto this source-derived direction:
\begin{equation}
S_{\mathrm{pro}}(z)=(z-\mu_0)^\top v_{\mathrm{norm}}, \quad
S_{\mathrm{pro}}(z)\in\mathbb{R}.
\end{equation}
The projection score measures the signed displacement of the test representation $z$ from the source cover mean along the cover--stego direction. We then obtain the edited representation $\tilde{z}$ by
\begin{equation}
\tilde{z}=z+\lambda S_{\mathrm{pro}}(z)v_{\mathrm{norm}},
\end{equation}
where $\lambda$ controls the editing strength. The edited representation $\tilde{z}$ is classified by the frozen classifier as in Eq.~(4).

This editing is sample-specific because $S_{\mathrm{pro}}(z)$ is computed for each test representation.
Unlike the DA vector, which applies a fixed translation to target samples, the DG editing is modulated by each sample's projection on the source-derived cover--stego direction.
Since different steganographic algorithms may induce representation shifts with different magnitudes, we use an algorithm-level coefficient $\lambda$ in the experiments.
For each steganographic algorithm, the same coefficient is fixed across all source--target transfer directions.

\section{Experiments}

\subsection{Basic Experimental Setup}

\textbf{Datasets and steganographic methods.}
We evaluate our method on three text corpora: Twitter \citep{go2009twitter}, Movie \citep{maas2011learning}, and News \citep{thompson2017allnews}. Five linguistic steganographic methods are considered, including Arithmetic Coding (AC) \citep{ziegler2019neural}, Huffman Coding (HC) \citep{yang2018automatically}, Adaptive Dynamic Grouping (ADG) \citep{zhang2021provably}, Meteor \citep{kaptchuk2021meteor}, and iMEC \citep{witt2023perfectly}. For each steganographic method, we construct six directed cross-corpus steganalysis tasks among the three corpora. Following the notation used in cross-domain steganalysis experiments, T, M, and N denote Twitter, Movie, and News, respectively. Thus, T-M, T-N, N-M, M-T, N-T, and M-N denote Twitter$\rightarrow$Movie, Twitter$\rightarrow$News, News$\rightarrow$Movie, Movie$\rightarrow$Twitter, News$\rightarrow$Twitter, and Movie$\rightarrow$News, respectively. For AC, HC, and ADG, stego texts are generated using LSTM language models trained on each corpus. 
For Meteor and iMEC, we use GPT-2 models \citep{radford2019language} fine-tuned on the corresponding corpus for stego-text generation. The average length (AL) and bits per word (BPW) of the generated texts are reported in Table~\ref{tab:data_statistics}.

\begin{table}[t]
\centering
\small
\caption{Statistics of Cover and Stego Text Datasets for Different Domains and Steganography Algorithms}
\label{tab:data_statistics}
\resizebox{\linewidth}{!}{
\begin{tabular}{lcccccc}
\toprule
\multirow{2}{*}{Method} 
& \multicolumn{2}{c}{News} 
& \multicolumn{2}{c}{Twitter} 
& \multicolumn{2}{c}{Movie} \\
\cmidrule(lr){2-3} \cmidrule(lr){4-5} \cmidrule(lr){6-7}
& AL & BPW & AL & BPW & AL & BPW \\
\midrule
Cover  & 19.13 & --   & 8.02  & --   & 18.94 & --   \\
AC     & 15.09 & 3.50 & 10.77 & 3.45 & 18.18 & 3.47 \\
HC     & 22.29 & 3.71 & 11.74 & 3.69 & 19.41 & 3.64 \\
ADG    & 22.75 & 5.41 & 10.90 & 5.13 & 20.66 & 5.24 \\
Meteor & 21.37 & 2.58 & 9.69  & 2.45 & 16.47 & 2.60 \\
iMEC   & 23.18 & 3.77 & 11.09 & 3.70 & 19.56 & 3.89 \\
\bottomrule
\end{tabular}
}
\end{table}

\textbf{Data construction.}
The cover texts are randomly sampled from the original corpora. Following previous cross-domain linguistic steganalysis studies, for each corpus and each steganographic method, the training set contains 10,000 cover texts and 10,000 stego texts. The validation set and the test set each contain 1,000 cover texts and 1,000 stego texts. The cover and stego samples are kept balanced in all splits. In addition, the training, validation, and test sets are strictly separated, and no text sample is allowed to appear in more than one split.

\textbf{Evaluation metrics.}
We use Accuracy (Acc) and F1-score (F1) as the evaluation metrics. The stego texts are treated as positive samples. The two metrics are defined as follows:
\begin{equation}
\mathrm{Acc} = \frac{TP + TN}{TP + TN + FP + FN},
\end{equation}
\begin{equation}
\mathrm{F1} = \frac{2TP}{2TP + FP + FN},
\end{equation}
where $TP$, $TN$, $FP$, and $FN$ denote true positives, true negatives, false positives, and false negatives, respectively.

\begin{table*}[!t]
\centering
\scriptsize
\setlength{\tabcolsep}{2.3pt}
\renewcommand{\arraystretch}{1.00}
\caption{Detailed Results on the Domain Adaptation Task}
\label{tab:da_results}
\resizebox{\textwidth}{!}{%
\begin{tabular}{llccccccc}
\toprule
Algorithm & Model 
& T-M & T-N & N-M & M-T & N-T & M-N & Avg. \\
\midrule
\multirow{4}{*}{AC}
& SANet        & 73.50/70.69 & 74.95/69.73 & \textbf{73.85}/\textbf{69.58} & 72.10/67.56 & 59.90/49.18 & 81.65/79.04 & 72.66/67.63 \\
& SANet+REED & \textbf{76.20}/\textbf{76.87} & \textbf{77.70}/\textbf{77.72} & 70.80/68.30 & \textbf{73.70}/71.96 & 60.05/58.92 & 82.15/81.78 & \textbf{73.43}/\textbf{72.59} \\
& CADA         & 65.65/67.86 & 58.95/61.14 & 68.85/65.97 & 65.50/68.31 & 55.78/\textbf{59.23} & \textbf{84.70}/\textbf{83.44} & 66.57/67.66 \\
& CADA+REED  & 73.35/73.80 & 76.15/75.77 & 63.60/56.24 & 71.70/\textbf{76.53} & \textbf{61.60}/54.72 & 79.20/80.97 & 70.93/69.67 \\
\midrule
\multirow{4}{*}{HC}
& SANet        & 74.80/75.75 & \textbf{66.10}/58.30 & \textbf{78.75}/77.19 & \textbf{74.20}/\textbf{74.78} & 72.50/69.72 & \textbf{72.00}/65.22 & \textbf{73.06}/70.16 \\
& SANet+REED & \textbf{76.70}/\textbf{78.33} & 64.85/\textbf{64.90} & \textbf{78.75}/\textbf{78.97} & 72.95/74.08 & 72.90/\textbf{73.06} & 67.80/\textbf{68.31} & 72.33/\textbf{72.94} \\
& CADA         & 72.90/72.43 & 64.20/55.25 & 76.25/73.03 & 69.45/64.37 & 69.15/61.37 & 66.20/53.69 & 69.69/63.36 \\
& CADA+REED  & 74.00/74.93 & 64.10/63.85 & 78.50/78.63 & 71.70/72.44 & \textbf{72.95}/72.80 & 68.00/67.81 & 71.54/71.74 \\
\midrule
\multirow{4}{*}{ADG}
& SANet        & 57.25/57.23 & 56.55/56.36 & 65.25/65.24 & 61.37/60.65 & 61.13/59.62 & 64.70/64.62 & 61.04/60.62 \\
& SANet+REED & 62.85/62.33 & \textbf{62.90}/\textbf{60.36} & \textbf{72.75}/\textbf{72.38} & \textbf{64.30}/\textbf{62.72} & \textbf{65.25}/58.31 & \textbf{73.05}/\textbf{70.66} & \textbf{66.85}/\textbf{64.46} \\
& CADA         & 52.60/52.59 & 54.36/54.31 & 52.31/52.15 & 53.39/53.35 & 52.95/52.69 & 53.74/53.67 & 53.23/53.13 \\
& CADA+REED  & \textbf{66.40}/\textbf{63.48} & 62.25/58.95 & 69.55/67.13 & 62.60/58.86 & 64.90/\textbf{63.40} & 69.35/65.43 & 65.84/62.88 \\
\midrule
\multirow{4}{*}{Meteor}
& SANet        & 74.55/77.43 & 66.45/67.47 & 77.20/78.02 & \textbf{77.55}/77.49 & \textbf{73.30}/71.20 & \textbf{79.50}/78.67 & 74.76/75.05 \\
& SANet+REED & \textbf{77.70}/\textbf{79.10} & \textbf{68.45}/\textbf{69.95} & \textbf{79.10}/\textbf{80.11} & 75.20/77.56 & 72.50/72.69 & 77.50/\textbf{79.24} & \textbf{75.08}/\textbf{76.44} \\
& CADA         & 65.32/66.28 & 60.05/63.23 & 69.10/72.06 & 64.50/67.42 & 68.40/72.49 & 69.80/73.57 & 66.19/69.17 \\
& CADA+REED  & 74.05/76.74 & 62.35/65.16 & 76.60/76.56 & 76.00/\textbf{78.59} & 72.20/\textbf{73.14} & 76.60/77.61 & 72.97/74.63 \\
\midrule
\multirow{4}{*}{iMEC}
& SANet        & 59.00/\textbf{63.35} & 57.75/53.59 & 67.75/67.74 & \textbf{71.20}/\textbf{73.43} & 66.70/65.89 & 67.60/72.11 & 65.00/66.02 \\
& SANet+REED & \textbf{61.85}/58.10 & \textbf{59.35}/52.92 & \textbf{70.65}/\textbf{79.22} & 69.80/67.56 & 66.25/65.61 & \textbf{73.95}/72.73 & \textbf{66.98}/66.02 \\
& CADA         & 58.37/57.08 & 56.27/53.49 & 66.64/71.25 & 70.34/68.45 & 65.36/63.40 & 65.53/70.89 & 63.75/64.09 \\
& CADA+REED  & 57.70/61.93 & 58.49/\textbf{53.72} & \textbf{70.65}/70.52 & 68.70/70.21 & \textbf{68.40}/\textbf{69.20} & 72.65/\textbf{73.64} & 66.10/\textbf{66.54} \\
\bottomrule
\end{tabular}
}
\end{table*}

\textbf{Implementation details.}
For the Domain Adaptation experiments, we follow the training hyperparameters of the corresponding backbone methods, SANet \citep{xue2024sanet} and CADA \citep{yang2025cada}. Following the balanced benchmark protocol used in previous cross-domain linguistic steganalysis studies, the domain-offset vector is estimated from 1,000 source-domain training samples and 1,000 target-domain samples. The target-domain samples are used to estimate only the marginal target-domain representation mean, rather than class-conditional target statistics. No target-domain labels are used by the proposed editing algorithm for classification loss, pseudo-label generation, class-conditional estimation, class-conditional optimization, or parameter updates. The vector coefficient is fixed to 1.

For the Domain Generalization experiments, we fully fine-tune BERT \citep{devlin2019bert} and the classifier on the labeled source-domain training set with a learning rate of $2 \times 10^{-5}$ for 3 epochs. The steganographic direction vector is estimated from 2,000 source-domain samples, including 1,000 cover texts and 1,000 stego texts. Since different steganographic algorithms may induce representation shifts with different magnitudes, we conduct an algorithm-level coefficient search for $\lambda$ within a predefined range. For each steganographic algorithm, the selected coefficient is fixed across all six source--target transfer directions, rather than tuned separately for each transfer task. The detailed coefficients are reported in Appendix~\ref{app:lambda}. 

For comparison with a parameter-update-based test-time adaptation baseline under the same source-trained setting, we apply a TTALS-style entropy-minimization update \citep{wu2024ttals} to the source-trained BERT-FT model, denoted as BERT-FT+TTem. This differs from the original TTALS setting, which performs fully test-time adaptation from a pre-trained language model without source-domain training. For BERT-FT+TTem, we use a learning rate of $1 \times 10^{-5}$ and perform one entropy-minimization step \citep{wang2021tent} for each target-domain batch.

\subsection{Results on Domain Adaptation}

We first evaluate the proposed method under the Domain Adaptation task by applying it to two representative cross-domain steganalysis methods, SANet \citep{xue2024sanet} and CADA \citep{yang2025cada}. For a fair comparison, we keep the same feature extractor, classifier, and training strategy as each backbone method. After the backbone model is trained, we compute the domain-offset vector from the marginal source and target representation means and apply it to the learned representations during testing. Table~\ref{tab:da_results} reports the Acc and F1 results on six directed cross-corpus tasks, with the last column showing the average performance over all tasks.

As shown in Table~\ref{tab:da_results}, adding the domain-offset vector improves the overall performance of both SANet and CADA on most cross-domain tasks. For SANet, the average Acc increases from 69.30\% to 70.93\%, and the average F1 increases from 67.90\% to 70.49\%. For CADA, the improvement is more pronounced, with the average Acc increasing from 63.89\% to 69.48\% and the average F1 increasing from 63.48\% to 69.09\%. These results indicate that our method is not tied to a specific backbone, but can be used as a post-training representation editing strategy for different cross-domain steganalysis models. The improvement on ADG is particularly notable. CADA+REED improves the average Acc/F1 from 53.23\%/53.13\% to 65.84\%/62.88\%, suggesting that when the original backbone still suffers from a clear source-target representation mismatch, the proposed vector editing can effectively compensate for such mismatch in the feature space.

It is worth noting that our experiments are mainly conducted under relatively high embedding-rate settings. 
In such settings, steganographic traces are generally more detectable, but they may also become more entangled with corpus-specific generation patterns and textual distributions, which makes cross-domain alignment more challenging. 
Although SANet and CADA reduce domain discrepancy through training-time alignment, they need to balance domain alignment with cover--stego discrimination. 
When steganographic variations are strong, this joint optimization may still leave residual source--target offsets or weaken class separation. 
By contrast, our method performs post-training representation editing. 
After the backbone has learned discriminative steganographic features, the estimated domain-offset vector directly compensates for the remaining source--target displacement before classification. 
In this way, vector editing can reduce residual domain mismatch while better preserving the learned cover--stego structure, which helps explain its effectiveness under high embedding-rate settings.

\subsection{Ablation Study on the Editing Strength in Domain Adaptation}
\label{sec:da_alpha_ablation}

We further analyze the effect of the editing strength $\alpha$ in the Domain Adaptation setting.
Here, $\alpha$ controls the magnitude of the domain-offset vector applied to target-domain representations.
When $\alpha=0$, no vector editing is performed, while larger values apply stronger representation editing along the estimated source--target offset direction.

\begin{figure}[t]
\centering
\includegraphics[width=0.85\linewidth]{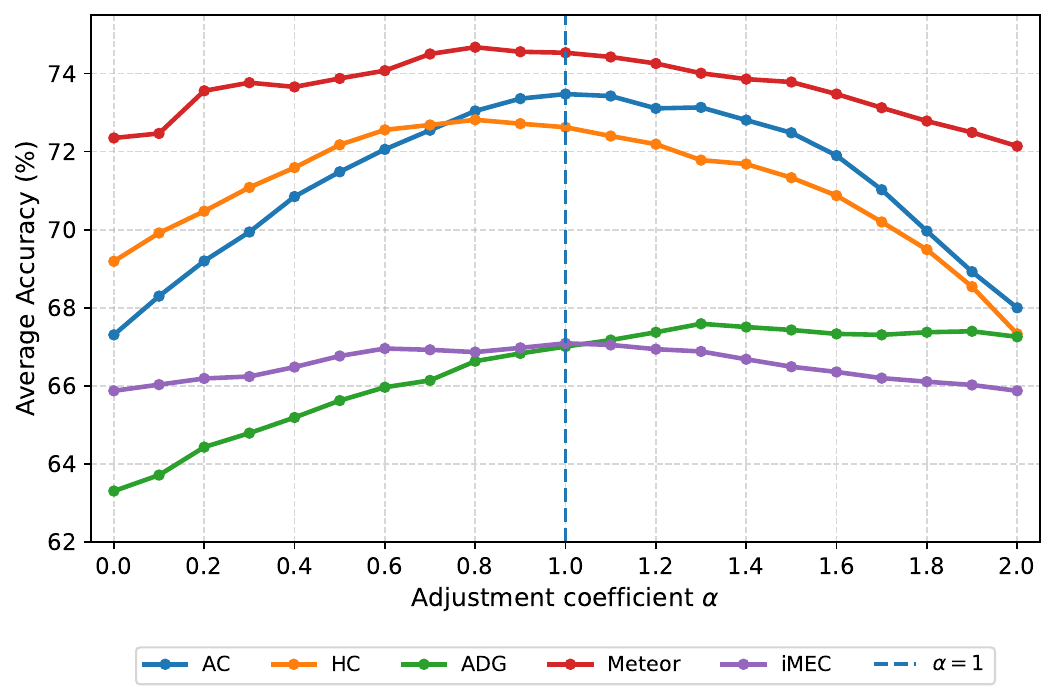}
\caption{Sensitivity of the editing strength $\alpha$ in domain adaptation.}
\label{fig:da_alpha_ablation}
\end{figure}

\begin{table*}[t]
\centering
\small
\setlength{\tabcolsep}{3.2pt}
\renewcommand{\arraystretch}{1.05}
\caption{Detailed Results on the Domain Generalization Task}
\label{tab:dg_results}
\resizebox{\textwidth}{!}{%
\begin{tabular}{llccccccc}
\toprule
Algorithm & Model & T-M & T-N & N-M & M-T & N-T & M-N & Avg. \\
\midrule

\multirow{3}{*}{AC}
& BERT-FT 
& 79.45/78.98 & 79.50/76.60 & 50.10/20.95 & 72.50/67.34 & 55.20/26.46 & 82.90/81.61 & 69.94/58.66 \\
& BERT-FT+TTem 
& \textbf{80.40}/80.06 & 61.60/39.14 & 50.95/3.91 & 53.10/13.79 & 50.95/4.26 & 82.50/79.82 & 63.25/36.83 \\
& BERT-FT+REED 
& 78.75/\textbf{81.64} & \textbf{81.95}/\textbf{82.31} & \textbf{62.95}/\textbf{47.93} & \textbf{77.10}/\textbf{79.80} & \textbf{63.70}/\textbf{58.94} & \textbf{84.35}/\textbf{84.22} & \textbf{74.80}/\textbf{72.47} \\

\midrule

\multirow{3}{*}{HC}
& BERT-FT 
& 77.70/76.05 & 64.15/50.59 & 74.40/67.43 & 71.75/65.14 & 67.00/53.59 & 69.10/59.32 & 70.68/62.02 \\
& BERT-FT+TTem 
& 64.20/73.04 & 52.65/12.74 & 51.65/6.75 & 55.00/18.77 & 50.95/4.11 & 54.45/16.65 & 54.82/22.01 \\
& BERT-FT+REED 
& \textbf{79.15}/\textbf{81.75} & \textbf{68.55}/\textbf{64.72} & \textbf{82.35}/\textbf{82.65} & \textbf{77.00}/\textbf{79.11} & \textbf{77.90}/\textbf{78.46} & \textbf{76.70}/\textbf{74.25} & \textbf{76.94}/\textbf{76.82} \\

\midrule

\multirow{3}{*}{ADG}
& BERT-FT 
& 68.30/62.97 & 65.75/57.37 & 73.30/68.33 & 66.50/60.77 & \textbf{67.45}/65.83 & 74.55/68.68 & 69.31/63.99 \\
& BERT-FT+TTem 
& 51.50/8.83 & 53.25/15.23 & 56.25/23.85 & \textbf{66.85}/70.01 & 57.85/69.93 & 54.25/16.29 & 56.66/34.02 \\
& BERT-FT+REED 
& \textbf{69.25}/\textbf{71.59} & \textbf{66.20}/\textbf{65.10} & \textbf{73.85}/\textbf{75.84} & 66.35/\textbf{72.22} & 63.45/\textbf{71.46} & \textbf{77.80}/\textbf{78.19} & \textbf{69.48}/\textbf{72.40} \\

\midrule

\multirow{3}{*}{Meteor}
& BERT-FT 
& \textbf{74.40}/\textbf{78.98} & 75.05/77.99 & \textbf{79.40}/\textbf{80.40} & \textbf{76.15}/75.01 & \textbf{75.15}/74.18 & \textbf{81.00}/79.30 & \textbf{76.86}/77.64 \\
& BERT-FT+TTem 
& 68.40/75.91 & 66.60/74.35 & 62.10/41.15 & 73.75/77.95 & 62.50/47.41 & 59.15/33.63 & 65.42/58.40 \\
& BERT-FT+REED 
& 69.95/74.50 & \textbf{77.65}/\textbf{80.52} & 73.80/78.24 & 75.40/\textbf{79.67} & 70.20/\textbf{76.06} & 76.95/\textbf{80.54} & 73.99/\textbf{78.26} \\

\midrule

\multirow{3}{*}{iMEC}
& BERT-FT 
& 55.89/67.23 & 60.78/65.45 & \textbf{69.89}/\textbf{65.41} & \textbf{70.02}/67.96 & \textbf{71.28}/65.92 & \textbf{73.19}/\textbf{75.98} & \textbf{66.84}/\textbf{67.99} \\
& BERT-FT+TTem 
& 49.76/50.66 & 50.19/53.41 & 57.56/55.74 & 65.13/58.26 & 50.29/55.13 & 67.98/60.58 & 56.82/55.63 \\
& BERT-FT+REED 
& \textbf{56.07}/\textbf{68.93} & \textbf{62.13}/\textbf{71.97} & 64.91/57.73 & 68.23/\textbf{69.46} & 60.98/\textbf{68.17} & 72.40/64.87 & 64.12/66.86 \\

\bottomrule
\end{tabular}
}
\end{table*}

As shown in Figure~\ref{fig:da_alpha_ablation}, the average accuracy generally increases as $\alpha$ grows from 0 to around 1, indicating that the estimated domain-offset vector effectively reduces the representation mismatch between the source and target domains. However, further increasing $\alpha$ does not consistently bring additional gains and may even lead to performance degradation. This suggests that overly strong editing may move target representations beyond the appropriate editing range.

Overall, setting $\alpha=1$ provides a stable and effective choice across different steganographic algorithms. This observation suggests that the estimated domain-offset vector already captures not only the editing direction but also a reasonable scale of the source--target displacement in the representation space. Therefore, our method does not require heavy tuning of the editing strength, and we fix $\alpha=1$ in all Domain Adaptation experiments.

\subsection{Results on Domain Generalization}

We further evaluate the proposed method under the Domain Generalization task. 
In this setting, we first fully fine-tune a BERT-based classifier \citep{devlin2019bert} on the labeled source domain. 
To ensure a fair comparison, all methods start from the same source-trained BERT-FT model. 
The source-trained model is directly used as the source-only BERT-FT baseline, while BERT-FT+TTem applies the entropy-minimization update of TTALS \citep{wu2024ttals} to this model. 
This setting differs from the original fully test-time adaptation setting of TTALS, but allows us to compare parameter-update-based test-time adaptation and our post-training representation editing under the same source-trained initialization. 
Table~\ref{tab:dg_results} reports the Acc and F1 results on six directed cross-corpus tasks, with the last column showing the average performance over all tasks.

As shown in Table~\ref{tab:dg_results}, our method improves the source-only BERT-FT baseline on most steganographic algorithms, especially in terms of F1-score. Averaged over all evaluated algorithms and transfer directions, our method improves Acc from 70.73\% to 71.87\%, and F1 from 66.06\% to 73.36\%. The improvement in F1 indicates that the source-derived steganographic direction helps reduce the imbalance between cover and stego detection on unseen target domains. The gains are especially notable on AC and HC. For AC, the average Acc/F1 increases from 69.94\%/58.66\% to 74.80\%/72.47\%. For HC, the average Acc/F1 increases from 70.68\%/62.02\% to 76.94\%/76.82\%. These results suggest that, once the editing coefficient is fixed, the cover--stego direction estimated from the source domain can provide useful guidance for target-domain representation editing.

Compared with TTALS, our method avoids test-time parameter updates and therefore shows more stable behavior in many tasks. TTALS degrades performance on several transfer directions, especially in terms of F1. This may be because entropy minimization updates model parameters according to the model's own predictions on target-domain samples. When the source-trained model produces biased or unreliable predictions, the entropy objective may reinforce incorrect decision tendencies and lead to imbalanced cover--stego detection. In contrast, our method keeps the source-trained model frozen and only edits the intermediate representation of each test sample along the source-derived cover--stego direction. This makes the editing process lightweight and less sensitive to noisy target predictions.

However, the improvement is not uniform across all steganographic algorithms. 
For iMEC, our method does not outperform the BERT-FT baseline on average, even with the selected algorithm-level coefficient. 
A possible reason is that iMEC places stronger emphasis on distributional security, aiming to make the generated stego texts distributionally close to cover texts. 
As a result, the cover--stego discrepancy induced by iMEC is usually weaker and more easily affected by corpus-specific topics, vocabularies, and generation patterns. 
In this case, the cover--stego direction estimated from the source domain may not mainly reflect a stable and transferable steganographic direction, but instead contain more source-specific discriminative cues. 
When this direction is applied to iMEC texts from unseen target domains, the projection score along this direction may fail to accurately indicate the strength of steganographic evidence, thereby limiting the effectiveness of representation editing along a single source-derived direction.

\section{Conclusion}

In cross-domain linguistic steganalysis, a detector trained on source-domain data often suffers from a mismatch between shifted target-domain representations and the decision boundary learned from the source domain. To address this problem, we proposed REED, a post-training representation editing framework. REED first trains a steganalysis classifier on labeled source-domain data and then keeps the feature extractor and classifier fixed. Instead of adapting the model by modifying parameters, adding training objectives, or moving the learned decision boundary, REED directly edits the intermediate representations of test samples before classification. This framework is instantiated under both domain adaptation and domain generalization settings: in domain adaptation, a source--target domain-offset vector is used to compensate for cross-domain shifts; in domain generalization, a source-derived cover--stego direction guides sample-specific editing for unseen-domain texts. Experiments on multiple corpora and steganographic algorithms show that REED improves cross-domain detection performance in most settings, especially in terms of F1-score, while requiring no architecture modification or parameter updates after source-domain training. These results indicate that feature-space representation editing provides a lightweight alternative to parameter-update-based adaptation. In future work, we will explore more robust vector estimation and automatic coefficient selection to further improve the adaptability of REED under more complex cross-domain scenarios.

\section{Limitation}

Although REED improves performance on most cross-domain linguistic steganalysis tasks, it still mainly relies on mean representations to construct editing vectors. This design is simple and efficient, but a single global vector may not fully capture fine-grained differences under complex distributions. When steganographic traces in the target domain differ substantially from those in the source domain, or are highly entangled with domain-specific topics and writing styles, the effectiveness of representation editing may be limited. The results on iMEC also show that the source-derived cover--stego direction does not always transfer stably to all target domains.

In addition, the current domain generalization experiments still require selecting the editing coefficient $\lambda$ at the steganographic-algorithm level. Although the same coefficient is fixed across all transfer directions for each algorithm, how to automatically determine an appropriate editing strength under a fully target-free setting remains an open problem. In future work, we will explore more robust vector estimation, adaptive direction selection, and automatic coefficient selection, and further evaluate REED on more text domains, steganographic algorithms, and pre-trained models.

\bibliography{custom}

@article{anderson1998limits,
  title={On The Limits of Steganography},
  author={Anderson, Ross J. and Petitcolas, Fabien A. P.},
  journal={IEEE Journal on Selected Areas in Communications},
  volume={16},
  number={4},
  pages={474--481},
  month={may},
  year={1998},
  doi={10.1109/49.668971},
  publisher={IEEE}
}

@inproceedings{ziegler2019neural,
  title={Neural Linguistic Steganography},
  author={Ziegler, Zachary M. and Deng, Yuntian and Rush, Alexander M.},
  booktitle={Proceedings of the 2019 Conference on Empirical Methods in Natural Language Processing and the 9th International Joint Conference on Natural Language Processing (EMNLP-IJCNLP)},
  pages={1210--1215},
  month={nov},
  year={2019},
  address={Hong Kong, China},
  publisher={Association for Computational Linguistics},
  doi={10.18653/v1/D19-1115},
  url={https://aclanthology.org/D19-1115/}
}

@inproceedings{zhang2021provably,
  title={Provably Secure Generative Linguistic Steganography},
  author={Zhang, Siyu and Yang, Zhongliang and Yang, Jinshuai and Huang, Yongfeng},
  booktitle={Findings of the Association for Computational Linguistics: ACL-IJCNLP 2021},
  pages={3046--3055},
  month={aug},
  year={2021},
  address={Online},
  publisher={Association for Computational Linguistics},
  doi={10.18653/v1/2021.findings-acl.268},
  url={https://aclanthology.org/2021.findings-acl.268/}
}

@article{zhou2022linguistic,
  title={Linguistic Steganography Based on Adaptive Probability Distribution},
  author={Zhou, Xuejing and Peng, Wanli and Yang, Boya and Wen, Juan and Xue, Yiming and Zhong, Ping},
  journal={IEEE Transactions on Dependable and Secure Computing},
  volume={19},
  number={5},
  pages={2982--2997},
  year={2022},
  doi={10.1109/TDSC.2021.3079957}
}

@article{mazurczyk2018information,
  title={Information Hiding: Challenges for Forensic Experts},
  author={Mazurczyk, Wojciech and Wendzel, Steffen},
  journal={Communications of the ACM},
  volume={61},
  number={1},
  pages={86--94},
  year={2018},
  doi={10.1145/3158416},
  publisher={Association for Computing Machinery}
}

@article{wen2019cnn,
  title={Convolutional Neural Network Based Text Steganalysis},
  author={Wen, Juan and Zhou, Xuejing and Zhong, Ping and Xue, Yiming},
  journal={IEEE Signal Processing Letters},
  volume={26},
  number={3},
  pages={460--464},
  month={mar},
  year={2019},
  doi={10.1109/LSP.2019.2895286},
  publisher={IEEE}
}

@article{yang2019tsrnn,
  title={{TS-RNN}: Text Steganalysis Based on Recurrent Neural Networks},
  author={Yang, Zhongliang and Wang, Ke and Li, Jian and Huang, Yongfeng and Zhang, Yu-Jin},
  journal={IEEE Signal Processing Letters},
  volume={26},
  number={12},
  pages={1743--1747},
  month={dec},
  year={2019},
  doi={10.1109/LSP.2019.2920452},
  publisher={IEEE}
}

@article{niu2019hybrid,
  title={A Hybrid {R-BILSTM-C} Neural Network Based Text Steganalysis},
  author={Niu, Yan and Wen, Juan and Zhong, Ping and Xue, Yiming},
  journal={IEEE Signal Processing Letters},
  volume={26},
  number={12},
  pages={1907--1911},
  month={dec},
  year={2019},
  doi={10.1109/LSP.2019.2953953},
  publisher={IEEE}
}

@article{peng2021realtime,
  title={{Real-Time Text Steganalysis Based on Multi-Stage Transfer Learning}},
  author={Peng, Wanli and Zhang, Jinyu and Xue, Yiming and Yang, Zhenghong},
  journal={IEEE Signal Processing Letters},
  volume={28},
  pages={1510--1514},
  year={2021},
  doi={10.1109/LSP.2021.3097241},
  publisher={IEEE}
}

@inproceedings{xue2022mda,
  title={{Domain Adaptational Text Steganalysis Based on Transductive Learning}},
  author={Xue, Yiming and Yang, Boya and Deng, Yaqian and Peng, Wanli and Wen, Juan},
  booktitle={Proceedings of the 2022 ACM Workshop on Information Hiding and Multimedia Security},
  pages={91--96},
  year={2022},
  doi={10.1145/3531536.3532963},
  publisher={ACM}
}

@article{xue2024sanet,
  title={{Adaptive Domain-Invariant Feature Extraction for Cross-Domain Linguistic Steganalysis}},
  author={Xue, Yiming and Wu, Jiaxuan and Ji, Ronghua and Zhong, Ping and Wen, Juan and Peng, Wanli},
  journal={IEEE Transactions on Information Forensics and Security},
  volume={19},
  pages={920--933},
  year={2024},
  doi={10.1109/TIFS.2023.3328455},
  publisher={IEEE}
}

@inproceedings{luo2025pdts,
  title={{Pseudo-label Based Domain Adaptation for Zero-Shot Text Steganalysis}},
  author={Luo, Yufei and Yang, Zhen and Zhang, Ru and Liu, Jianyi},
  booktitle={Computational and Experimental Simulations in Engineering},
  series={Mechanisms and Machine Science},
  volume={176},
  pages={128--142},
  year={2025},
  editor={Zhou, Kun},
  publisher={Springer},
  address={Cham},
  doi={10.1007/978-3-031-82907-9_10}
}

@article{yang2025cada,
  title={{Class-Aware Adversarial Unsupervised Domain Adaptation for Linguistic Steganalysis}},
  author={Yang, Zhen and Luo, Yufei and Yang, Jinshuai and Xu, Xin and Zhang, Ru and Huang, Yongfeng},
  journal={IEEE Transactions on Information Forensics and Security},
  volume={20},
  pages={5181--5194},
  year={2025},
  doi={10.1109/TIFS.2025.3569409},
  publisher={IEEE}
}

@article{wu2024ttals,
  title={{A Test-Time Entropy Minimization Method for Cross-Domain Linguistic Steganalysis}},
  author={Wu, Jiaxuan and Chen, Xin and Wen, Juan and Peng, Wanli and Xue, Yiming},
  journal={IEEE Signal Processing Letters},
  volume={31},
  pages={1760--1764},
  year={2024},
  doi={10.1109/LSP.2024.3421946},
  publisher={IEEE}
}

@inproceedings{wang2021tent,
  title={{Tent: Fully Test-Time Adaptation by Entropy Minimization}},
  author={Wang, Dequan and Shelhamer, Evan and Liu, Shaoteng and Olshausen, Bruno and Darrell, Trevor},
  booktitle={International Conference on Learning Representations},
  year={2021},
  url={https://openreview.net/forum?id=uXl3bZLkr3c}
}

@inproceedings{qi2013secure,
  title={{A Secure Text Steganography Based on Synonym Substitution}},
  author={Cao, Qi and Sun, Xingming and Xiang, Lingyun},
  booktitle={IEEE Conference Anthology},
  pages={1--3},
  year={2013},
  doi={10.1109/ANTHOLOGY.2013.6784896},
  publisher={IEEE}
}

@article{xiang2014linguistic,
  title={{Linguistic Steganalysis Using the Features Derived from Synonym Frequency}},
  author={Xiang, Lingyun and Sun, Xingming and Luo, Gang and Xia, Bin},
  journal={Multimedia Tools and Applications},
  volume={71},
  pages={1893--1911},
  year={2014},
  doi={10.1007/s11042-012-1313-8},
  publisher={Springer}
}

@inproceedings{shirali2008text,
  title={{Text Steganography by Changing Words Spelling}},
  author={Shirali-Shahreza, Mohammad},
  booktitle={2008 10th International Conference on Advanced Communication Technology},
  volume={3},
  pages={1912--1913},
  year={2008},
  doi={10.1109/ICACT.2008.4494159},
  publisher={IEEE}
}

@inproceedings{fang2017generating,
  title={{Generating Steganographic Text with {LSTM}s}},
  author={Fang, Tina and Jaggi, Martin and Argyraki, Katerina},
  booktitle={Proceedings of {ACL} 2017, Student Research Workshop},
  pages={100--106},
  year={2017},
  month={jul},
  address={Vancouver, Canada},
  publisher={Association for Computational Linguistics},
  url={https://aclanthology.org/P17-3017/}
}

@article{yang2019rnnstega,
  title={{RNN-Stega}: Linguistic Steganography Based on Recurrent Neural Networks},
  author={Yang, Zhong-Liang and Guo, Xiao-Qing and Chen, Zi-Ming and Huang, Yong-Feng and Zhang, Yu-Jin},
  journal={IEEE Transactions on Information Forensics and Security},
  volume={14},
  number={5},
  pages={1280--1295},
  year={2019},
  doi={10.1109/TIFS.2018.2871746},
  publisher={IEEE}
}

@misc{yang2018automatically,
  title={{Automatically Generate Steganographic Text Based on Markov Model and Huffman Coding}},
  author={Yang, Zhongliang and Jin, Shuyu and Huang, Yongfeng and Zhang, Yujin and Li, Hui},
  year={2018},
  eprint={1811.04720},
  archivePrefix={arXiv},
  primaryClass={cs.CR},
  doi={10.48550/arXiv.1811.04720},
  url={https://arxiv.org/abs/1811.04720}
}

@inproceedings{kaptchuk2021meteor,
  title={{Meteor}: Cryptographically Secure Steganography for Realistic Distributions},
  author={Kaptchuk, Gabriel and Jois, Tushar M. and Green, Matthew and Rubin, Aviel D.},
  booktitle={Proceedings of the 2021 ACM SIGSAC Conference on Computer and Communications Security},
  pages={1529--1548},
  year={2021},
  doi={10.1145/3460120.3484550},
  publisher={ACM}
}

@inproceedings{witt2023perfectly,
  title={{Perfectly Secure Steganography Using Minimum Entropy Coupling}},
  author={Schroeder de Witt, Christian and Sokota, Samuel and Kolter, J. Zico and Foerster, Jakob Nicolaus and Strohmeier, Martin},
  booktitle={International Conference on Learning Representations},
  year={2023},
  url={https://openreview.net/forum?id=HQ67mj5rJdR}
}

@inproceedings{devlin2019bert,
  title={{BERT}: Pre-training of Deep Bidirectional Transformers for Language Understanding},
  author={Devlin, Jacob and Chang, Ming-Wei and Lee, Kenton and Toutanova, Kristina},
  booktitle={Proceedings of the 2019 Conference of the North American Chapter of the Association for Computational Linguistics: Human Language Technologies, Volume 1 (Long and Short Papers)},
  pages={4171--4186},
  year={2019},
  month={jun},
  address={Minneapolis, Minnesota},
  publisher={Association for Computational Linguistics},
  doi={10.18653/v1/N19-1423},
  url={https://aclanthology.org/N19-1423/}
}

@article{gretton2012kernel,
  title={{A Kernel Two-Sample Test}},
  author={Gretton, Arthur and Borgwardt, Karsten M. and Rasch, Malte J. and Sch{\"o}lkopf, Bernhard and Smola, Alexander},
  journal={Journal of Machine Learning Research},
  volume={13},
  number={25},
  pages={723--773},
  year={2012}
}

@misc{go2009twitter,
  title={{Twitter Sentiment Classification Using Distant Supervision}},
  author={Go, Alec and Bhayani, Richa and Huang, Lei},
  year={2009},
  note={{CS224N} Project Report, Stanford University},
  url={https://www-cs.stanford.edu/people/alecmgo/papers/TwitterDistantSupervision09.pdf}
}

@inproceedings{maas2011learning,
  title={{Learning Word Vectors for Sentiment Analysis}},
  author={Maas, Andrew L. and Daly, Raymond E. and Pham, Peter T. and Huang, Dan and Ng, Andrew Y. and Potts, Christopher},
  booktitle={Proceedings of the 49th Annual Meeting of the Association for Computational Linguistics: Human Language Technologies},
  pages={142--150},
  year={2011},
  month={jun},
  address={Portland, Oregon, USA},
  publisher={Association for Computational Linguistics},
  url={https://aclanthology.org/P11-1015/}
}

@misc{thompson2017allnews,
  title={{All the News: 143,000 Articles from 15 American Publications}},
  author={Thompson, Andrew},
  year={2017},
  note={{Kaggle Dataset, Version 4}}
}

@techreport{radford2019language,
  title={Language Models are Unsupervised Multitask Learners},
  author={Radford, Alec and Wu, Jeffrey and Child, Rewon and Luan, David and Amodei, Dario and Sutskever, Ilya},
  institution={OpenAI},
  year={2019},
  url={https://cdn.openai.com/better-language-models/language_models_are_unsupervised_multitask_learners.pdf}
}

@article{wu2021linguistic,
  title={{Linguistic Steganalysis With Graph Neural Networks}},
  author={Wu, Hanzhou and Yi, Biao and Ding, Feng and Feng, Guorui and Zhang, Xinpeng},
  journal={IEEE Signal Processing Letters},
  volume={28},
  pages={558--562},
  year={2021},
  doi={10.1109/LSP.2021.3062233},
  publisher={IEEE}
}

@article{xue2022effective,
  title={{An Effective Linguistic Steganalysis Framework Based on Hierarchical Mutual Learning}},
  author={Xue, Yiming and Kong, Lingzhi and Peng, Wanli and Zhong, Ping and Wen, Juan},
  journal={Information Sciences},
  volume={586},
  pages={140--154},
  year={2022},
  doi={10.1016/j.ins.2021.11.086},
  publisher={Elsevier}
}

@article{yang2023linguistic,
  title={{Linguistic Steganalysis Toward Social Network}},
  author={Yang, Jinshuai and Yang, Zhongliang and Zou, Jiajun and Tu, Haoqin and Huang, Yongfeng},
  journal={IEEE Transactions on Information Forensics and Security},
  volume={18},
  pages={859--871},
  year={2023},
  doi={10.1109/TIFS.2022.3226909},
  publisher={IEEE}
}

@article{peng2023text,
  title={{Text Steganalysis Based on Hierarchical Supervised Learning and Dual Attention Mechanism}},
  author={Peng, Wanli and Li, Sheng and Qian, Zhenxing and Zhang, Xinpeng},
  journal={IEEE/ACM Transactions on Audio, Speech, and Language Processing},
  volume={31},
  pages={3513--3526},
  year={2023},
  doi={10.1109/TASLP.2023.3319975},
  publisher={IEEE}
}

@article{zhou2022domain,
  title={Domain Generalization: A Survey},
  author={Zhou, Kaiyang and Liu, Ziwei and Qiao, Yu and Xiang, Tao and Loy, Chen Change},
  journal={IEEE Transactions on Pattern Analysis and Machine Intelligence},
  volume={45},
  number={4},
  pages={4396--4415},
  year={2023},
  publisher={IEEE}
}

@inproceedings{sun2020test,
  title={{Test-Time Training With Self-Supervision for Generalization Under Distribution Shifts}},
  author={Sun, Yu and Wang, Xiaolong and Liu, Zhuang and Miller, John and Efros, Alexei and Hardt, Moritz},
  booktitle={Proceedings of the 37th International Conference on Machine Learning},
  series={Proceedings of Machine Learning Research},
  volume={119},
  pages={9229--9248},
  year={2020},
  publisher={PMLR}
}

@inproceedings{niu2022efficient,
  title={{Efficient Test-Time Model Adaptation Without Forgetting}},
  author={Niu, Shuaicheng and Wu, Jiaxiang and Zhang, Yifan and Chen, Yaofo and Zheng, Shijian and Zhao, Peilin and Tan, Mingkui},
  booktitle={Proceedings of the 39th International Conference on Machine Learning},
  series={Proceedings of Machine Learning Research},
  volume={162},
  pages={16888--16905},
  year={2022},
  publisher={PMLR}
}

\appendix
\appendix

\section{Editing Coefficients}
\label{app:lambda}

Table~\ref{tab:lambda} reports the algorithm-level editing coefficients selected from the predefined search range in the Domain Generalization experiments. 
Specifically, we search $\lambda$ in the range of $[0, 10]$ with a step size of $0.1$. 
For each steganographic algorithm, the same coefficient is used for all six source--target transfer directions. 
The coefficient $\lambda$ controls the strength of the steganographic direction editing.

\begin{table}[H]
\centering
\resizebox{\linewidth}{!}{%
\begin{tabular}{cccccc}
\toprule
Algorithm & AC & HC & ADG & Meteor & iMEC \\
\midrule
$\lambda$ & 5.0 & 2.4 & 0.4 & 0.2 & 0.1 \\
\bottomrule
\end{tabular}
}
\caption{Algorithm-level editing coefficients selected from the predefined search range in the Domain Generalization experiments.}
\label{tab:lambda}
\end{table}

\end{document}